\documentclass{article}

\usepackage[english]{babel}

\usepackage[letterpaper,top=2cm,bottom=2cm,left=3cm,right=3cm,marginparwidth=1.75cm]{geometry}

\usepackage{amsmath}
\usepackage{graphicx}
\usepackage{caption}
\usepackage{subcaption}
\usepackage{makecell} % allows line breaks inside table cells
\usepackage{array} 
\usepackage{authblk}
\usepackage[colorlinks=true, allcolors=blue]{hyperref}

\newcommand\blfootnote[1]{%
  \begingroup
  \renewcommand\thefootnote{}\footnote{#1}%
  \addtocounter{footnote}{-1}%
  \endgroup
}

\title{QuantX: A Framework for Hardware-Aware Quantization of Generative AI Workloads}

\author{Muhammad Ahmad}
\author{Khurram Mazher}
\author{Saqib Akram}
\author{Ahmad Tameem}
\author{\\Saad Bin Nasir}
\affil {10xEngineers Inc.}

\begin{document}
\maketitle

\blfootnote{The authors can be reached at ``khurram.usman@10xengineers.ai" and ``saad.nasir@10xengineers.ai" for any queries.}

%%%%%%%%%%%%%%%%%%%%%%%%%%%%%%%%%%%%%%%%%%%%%%%%%%%%%%%%%%%%%%%%%%%%%%%%%%%%%%%%%%%%%%%
\vspace{-1cm}
\begin{abstract}
We present QuantX: a tailored suite of recipes for LLM and VLM quantization. It is capable of quantizing down to 3-bit resolutions with minimal loss in performance. The quantization strategies in QuantX take into account hardware-specific constraints to achieve efficient dequantization during inference ensuring flexible trade-off between runtime speed, memory requirement and model accuracy. Our results demonstrate that QuantX achieves performance within 6\% of the unquantized model for LlaVa-v1.6 quantized down to 3-bits for multiple end user tasks and outperforms recently published state-of-the-art quantization techniques. We further integrate one particular technique from QuantX into the popular Llama.cpp framework and show its feasibility in terms of runtime compared to the mainstream quantization techniques from Llama.cpp. Lastly, this manuscript provides insights into the LLM quantization process that motivated the range of recipes and options that are incorporated in QuantX.
\end{abstract}

%%%%%%%%%%%%%%%%%%%%%%%%%%%%%%%%%%%%%%%%%%%%%%%%%%%%%%%%%%%%%%%%%%%%%%%%%%%%%%%%%%%%%%%
\section{Introduction}

With new Large/Visual Language Models (LLMs / VLMs) coming out with an increasing frequency and finding end-user applications, there has been a recent trend in enabling local inference of these LLMs particularly in the mobile and edge use-cases. Local inference is especially important for privacy-critical applications where the user does not want the information to be sent over the internet. Under limited memory and compute power, LLMs have to be compressed in size with minimal loss in performance. Compared to quantization-aware training, Post-Training Quantization (PTQ) is a relatively low cost method for compression and has been the topic of active research for the last few years \cite{SQ,AWQ,spinQuant,quip}. In PTQ, pretrained model weights are quantized to lower bit resolutions using a small calibration dataset. 

% \href{https://www.overleaf.com/learn}{help library}

%%%%%%%%%%%%%%%%%%%%%%%%%%%%%%%%%%%%%%%%%%%%%%%%%%%%%%%%%%%%%%%%%%%%%%%%%%%%%%%%%%%%%%%
\section{QuantX Framework}
In this section, we discuss a few interesting insights into the LLM quantization process. We also discuss a few design choices made in the various recipes available in the QuantX quantization framework as a result of these insights.

\subsection{Understanding PDF Differences across Models and Layers} \label{sec:PDF}
\begin{figure}[t]
     \centering
     \begin{subfigure}[t]{0.49\textwidth}
         \centering
         \includegraphics[width=\textwidth]{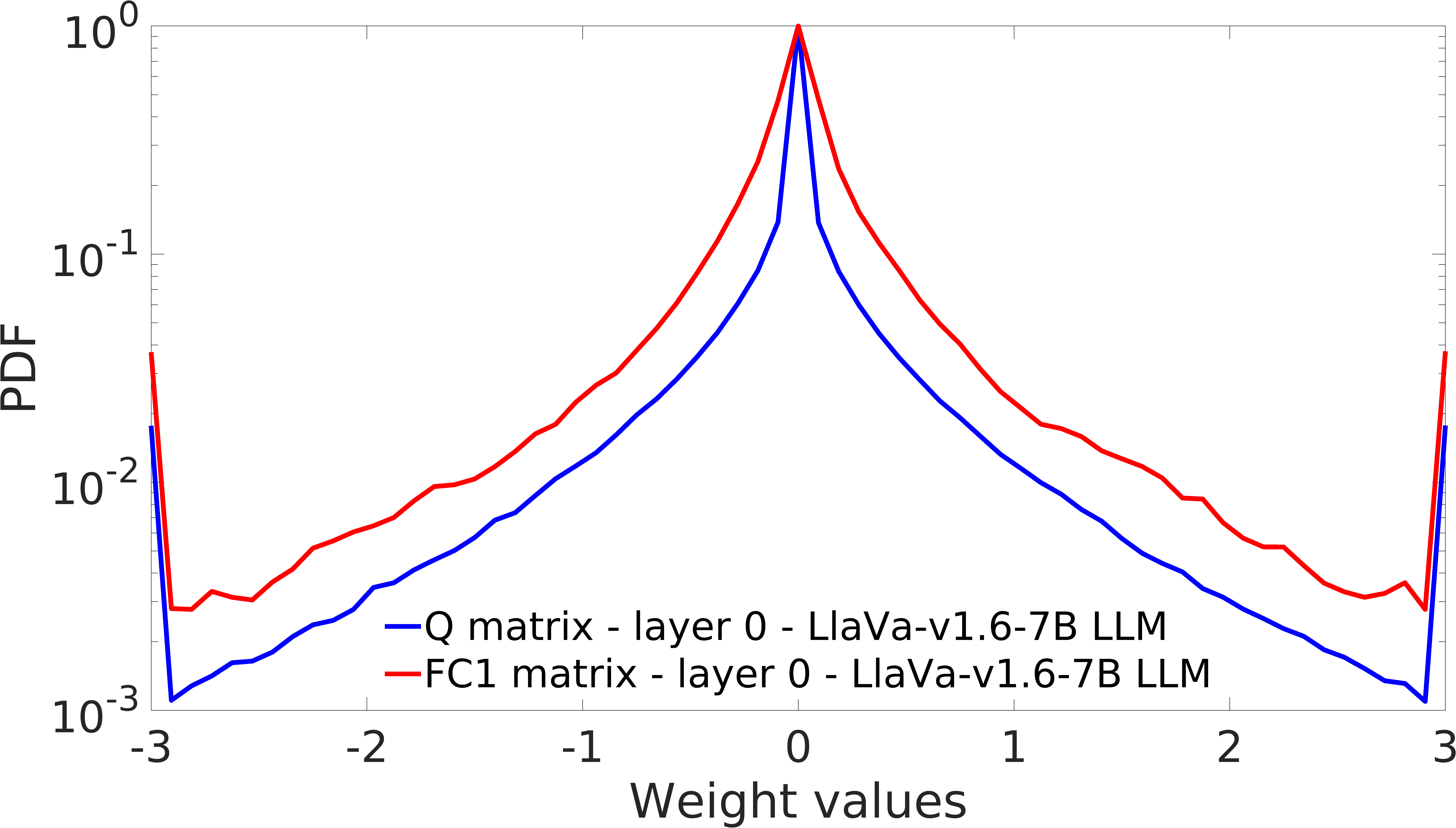}
         \caption{Normalized histograms of the weight matrix $\mathbf{Q}$ from the SA module (in \textcolor{blue}{blue}) and the weight matrix $\mathbf{FC1}$ from the MLP module (in \textcolor{red}{red}) of layer 0 of the Llava-v1.6 LLM.}
         \label{fig:pdfSameModel}
     \end{subfigure}
     \hfill
     \begin{subfigure}[t]{0.49\textwidth}
         \centering
         \includegraphics[width=\textwidth]{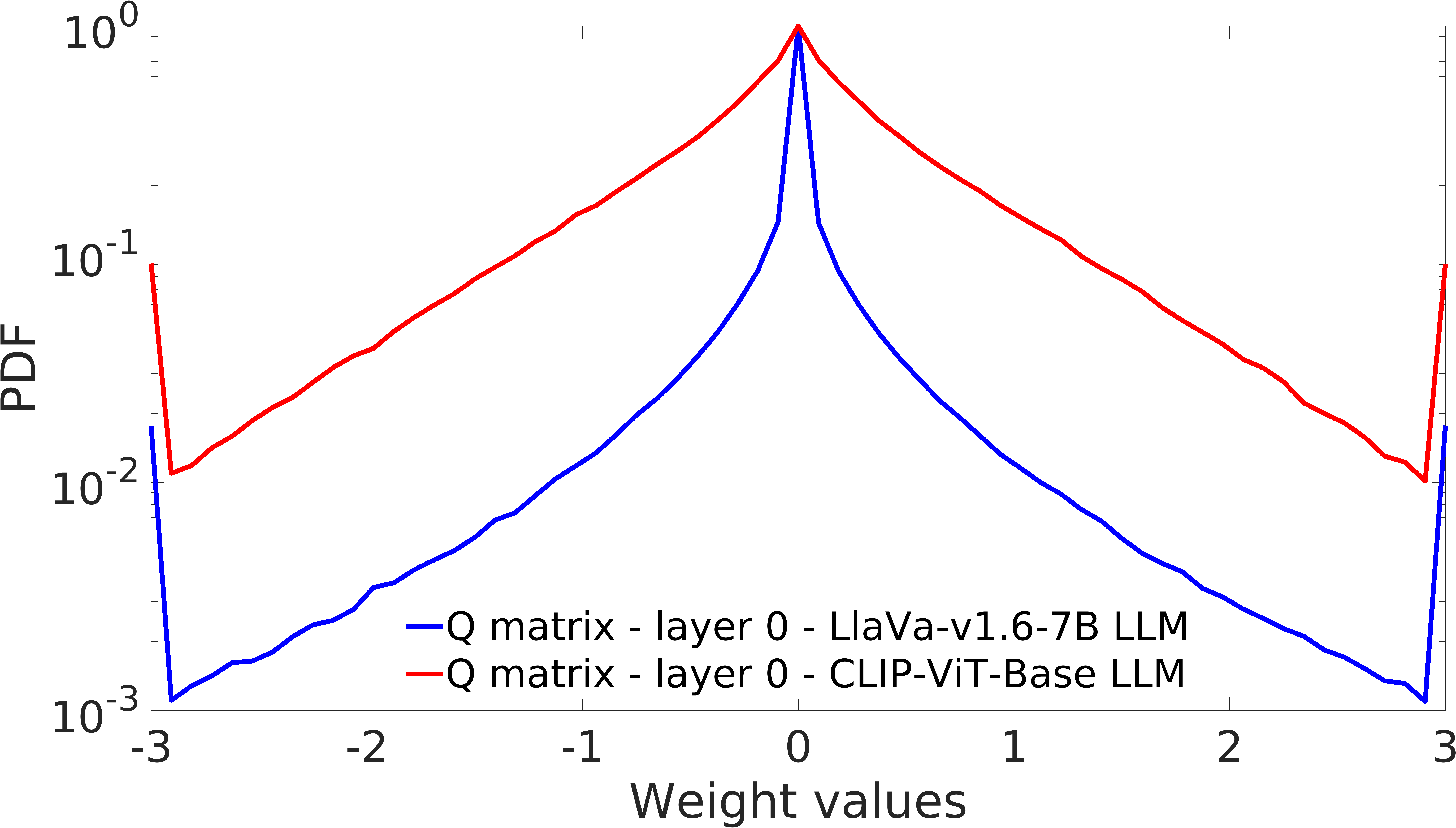}
         \caption{Normalized histograms of the weight matrix $\mathbf{Q}$ from the SA module of layer 0 of the LlaVa-v1.6 LLM (in \textcolor{blue}{blue}) and the weight matrix $\mathbf{Q}$ from the SA module of layer 0 of the CLIP-ViT-Base text encoder (in \textcolor{red}{red}).}
         \label{fig:pdfSameLayer}
     \end{subfigure}
        \caption{It can be observed there is significant difference in the shapes across different layers of the same models as well as same layers of the different models. The bit-width resolution and the choice between uniform and non-uniform quantization in QuantX can be tailored based on these statistics.}
        \label{fig:pdfSame}
        \vspace{-0.2cm}
\end{figure}

The distribution of the weight parameters across different layers of the same model differ significantly. The PDF of the weight matrix $\mathbf{Q}$ of the Self-Attention (SA) module and the weight matrix $\mathbf{FC1}$ of the Multi-Layer Perceptron (MLP) module from layer 0 of the language model of LlaVa-v1.6 \cite{llava} are shown in Fig. \ref{fig:pdfSameModel} with the y-axis plotted in logarithmic scale. It can be observed that the weights of the matrix $\mathbf{Q}$ have a smaller variance compared to the weights of the matrix $\mathbf{FC1}$ with significantly more of its weights centered around 0. This makes quantizing the $\mathbf{Q}$ matrix different compared to the matrix $\mathbf{FC1}$ which can be optimized by carefully choosing different bit-widths and between uniform and non-uniform quantization. We also note that the weight distribution of the same layers across different models can vary significantly. As shown in Fig. \ref{fig:pdfSameLayer}, the distribution of the weight matrix $\mathbf{Q}$ from the SA module of layer 0 has a larger variance for the CLIP model \cite{clip} compared to the LlaVa-v1.6 model \cite{llava}, making it more amenable to low-bit quantization.% through the use of non-uniform techniques.

% and under the presence of any outlier weights

\subsection{Critical Outliers} \label{sec:outliers}
Given the big variance of the weight distribution across different models and layers, it is expected that a non-uniform quantization strategy based on the PDF of the weights will outperform a uniform quantization strategy with equally spaced points. Interestingly, this is not always the case. For a group of points mostly distributed around 0 (as shown in Fig. \ref{fig:outliers} for a group from the $\mathbf{Q}$ matrix of layer 0 in LlaVa-v1.6), a non-uniform quantization strategy will place more points in the region where most of the probability mass is concentrated. A consequence of such a placement is that any weight parameters outside the region of high probability mass (highlighted in black circles in Fig. \ref{fig:outliers}) will have larger quantization errors, effectively acting as outliers. Note that our definition of an outlier weight here is different from what is typically published in the literature. Our learning suggests a large quantization error on these outlier weights could draw a significant drop in mobile-class LLM's accuracy. Drawing on this insight, QuantX incorporates both uniform and non-uniform quantization of the weight matrices and chooses between the two based on various statistics and constraints.

% \subsection{Attention map delta}
\subsection{Multi-Criterion Approach across Models and Layers}
\begin{figure}[t]
     \centering
     \begin{subfigure}[t]{0.49\textwidth}
         \centering
         \includegraphics[width=\textwidth]{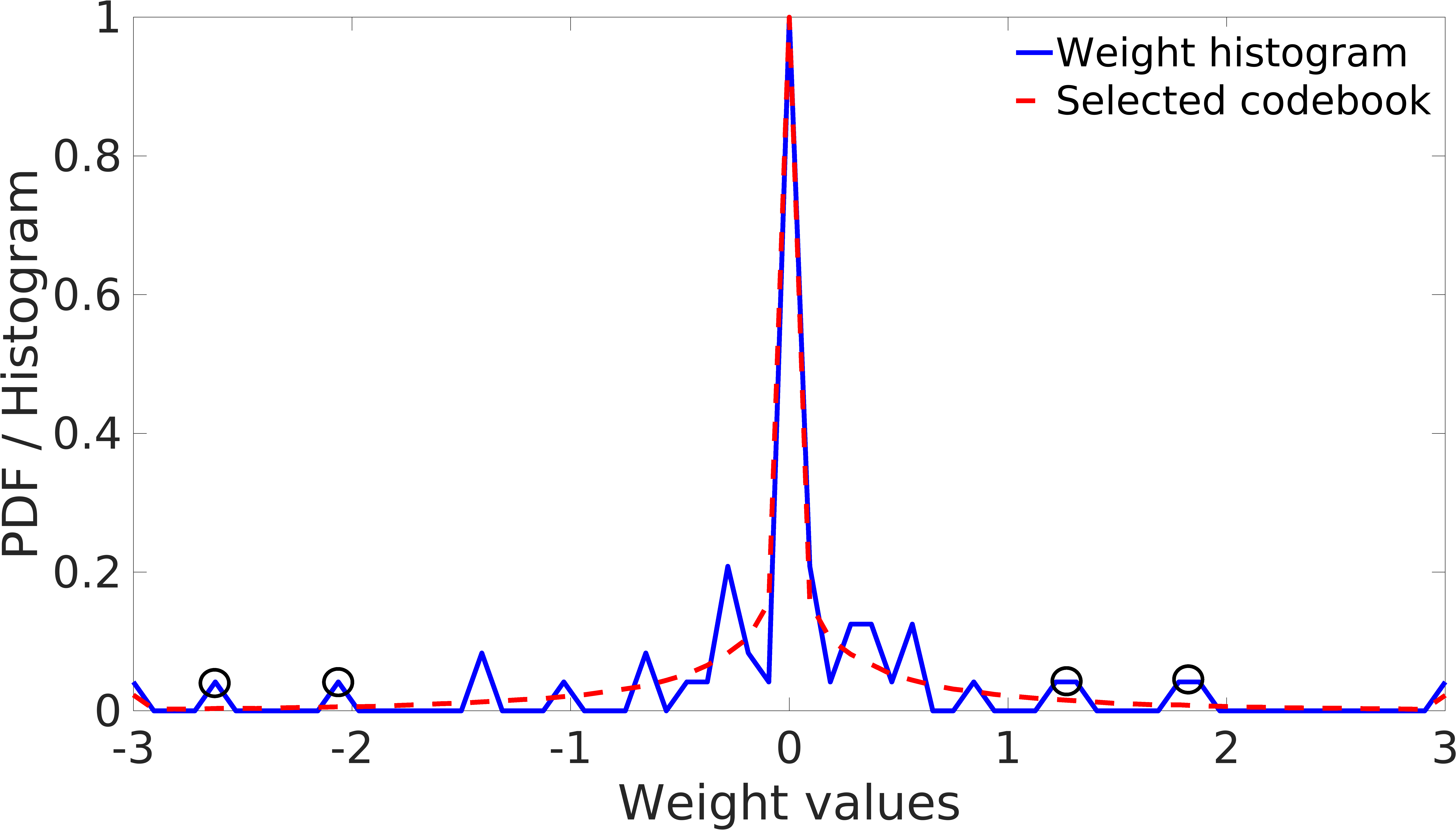}
         \caption{Normalized histogram of a particular group of weights from the weight matrix $\mathbf{Q}$ of the SA module of layer 0 of the LlaVa-v1.6 LLM (in \textcolor{blue}{blue}).  Shown in \textcolor{red}{red} is the codebook to which the same group was mapped by the non-uniform quantization recipe in QuantX. The black circles highlight the outlier weights in the region of the codebook with \emph{almost} zero probability mass resulting in large quantization errors.}
         \label{fig:outliers}
     \end{subfigure}
     \hfill
     \begin{subfigure}[t]{0.49\textwidth}
         \centering
         \includegraphics[width=\textwidth]{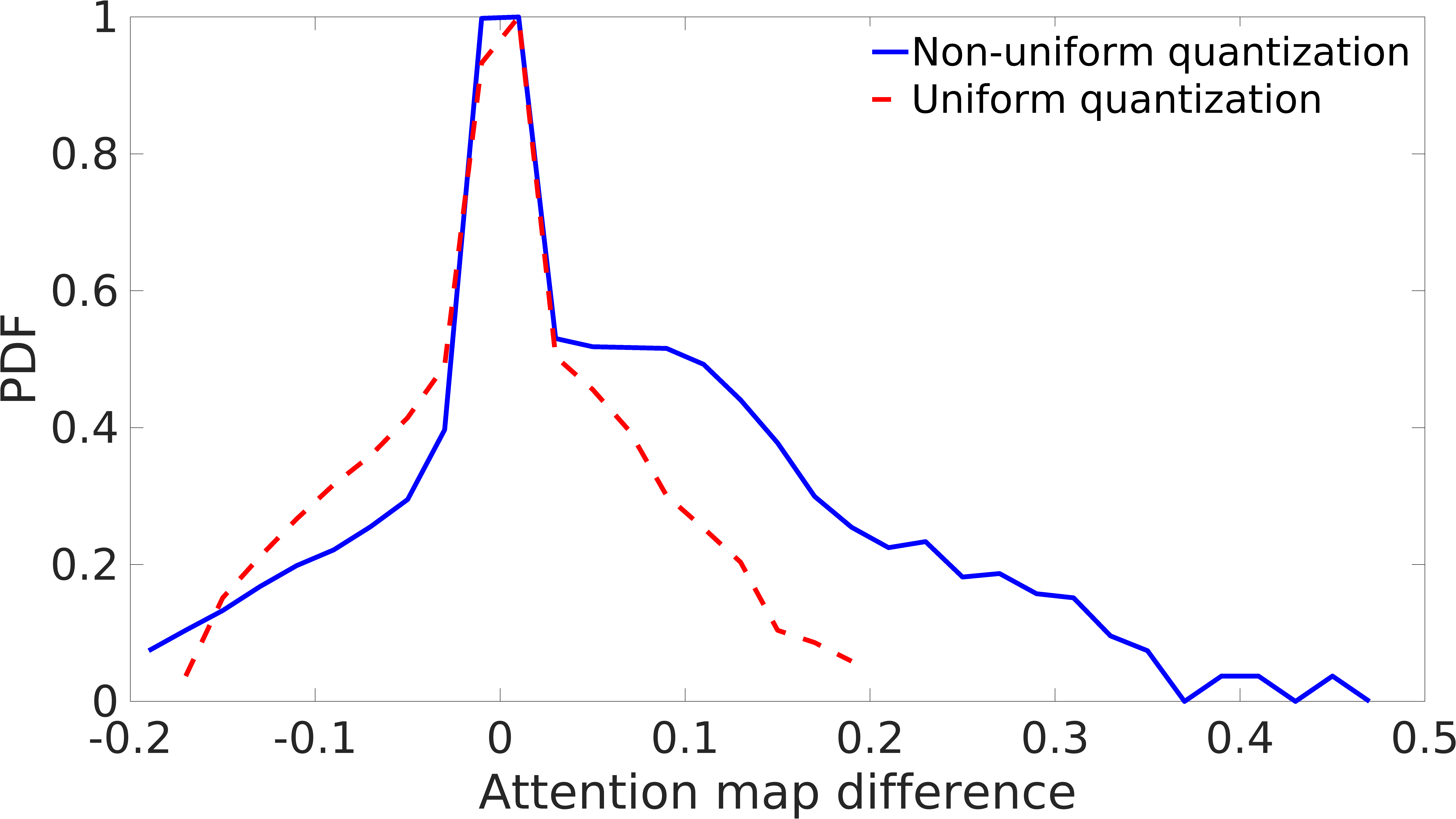}
         \caption{Normalized histograms of the attention map difference between the unquantized and quantized models from layer 0 of the LlaVa-v1.6 LLM. It can be seen that in terms of minimizing the token to token attention discrepancy, the uniform quantization (in \textcolor{red}{red}) performs better than the non-uniform one (in \textcolor{blue}{blue}). The choice between uniform and non-uniform quantization can be made based on this on a per-layer level.}
         \label{fig:delta}
     \end{subfigure}
        \caption{Illustration of the critical outliers problem and attention map distortion as a result of non-uniform quantization. }
        \label{fig:fig2}
        \vspace{-0.2cm}
\end{figure}

While an individual matrix of a particular layer (for example $\mathbf{Q}$ matrix of layer 0 in LlaVa-v1.6) may be better quantized by a non-uniform quantization approach evaluated on the Frobenius norm error compared to a uniform approach, it may not result in an equivalent end-to-end LLM accuracy recovery at low-bit resolutions. As such, other intermediate references, if available, must be utilized in addition to the error norm.  One such intermediate reference is the self-attention output (token to token importance) in transformer models. We demonstrate that the layer-wise attention maps are affected by the type of quantization process and differently so for the uniform and non-uniform quantization as shown in Fig. \ref{fig:delta}. Maintaining the same token-to-token importance in the attention map matrix takes precedence over quantizing the $\mathbf{K}$ and $\mathbf{Q}$ matrices with minimum MSE or similar regression metrics.  

Unlike the attention block where considerable information is stored within the tokens, the MLP distributes this information. Therefore, quantization settings would change accordingly. As demonstrated in Section \ref{sec:PDF}, the variance of the weight matrices of the MLP is larger compared to the self-attention block. This reduces the effect of any outlier weights discussed in Section \ref{sec:outliers}. All of these insights together make the MLP block a better candidate for non-uniform quantization methods. 

It is well known that different layers within a model could be quantized with different bit-widths to yield the best trade-off of memory, inference speed and accuracy \cite{hawq}. A similar trend follows at the multi-modal level, where different quantization techniques and bit-widths across sub-models yield better preservation of performance from an end-user task perspective.

\subsection{Hardware-Aware Constraints}
The various aspects of the quantization process of the weight matrices discussed so far did not account for the availability of hardware resources. In order to achieve effective inference speed and memory improvements for the quantized model, we have to ensure the QuantX recipes remain in the realm of deployable techniques and not overly complicated ones which can result in loss of inference speed. Supported numerics (INT4, INT8, MX, etc), types of multipliers, data-mover bus widths, etc can together change the Pareto frontier of the 3-dimensional space of inference speed, memory consumption and model accuracy. In most of the published work with significant industry traction, activations are typically quantized to INT8 values and not below. Activations, however, tend to have extreme outliers particularly for larger models \cite{SQ} making their quantization a difficult task. In such cases, migrating the quantization difficulty between the weights and the activations through a mathematically equivalent scaling like done by \cite{SQ,AWQ} becomes a must-have feature. 

The type and speed of software kernels that enable optimal usage of hardware features also feeds in deciding the optimal quantization strategy. LLM decoding is known to be a memory-bound operation and as such packing and unpacking kernels play an important role in determining the run-time.  Recent SOTA quantization techniques \cite{spinQuant,quip} have suggested the use of transformation operations before quantization which can reduce the loss due to quantization. Such transformations, however, have to be reversed during dequantization, which increases the runtime and reduces the achieved tokens per second. If maximizing inference speed is the main target, then such techniques may not be suitable. On the other hand, if accuracy in limited memory is desired, then custom dequantization kernels enable that. This, again, motivates the various hardware-informed design choices made in~QuantX.

%%%%%%%%%%%%%%%%%%%%%%%%%%%%%%%%%%%%%%%%%%%%%%%%%%%%%%%%%%%%%%%%%%%%%%%%%%%%%%%%%%%%%%%
\section{Results}
In this section, we present our performance and run-time results on the VLM LlaVa-v1.6 7B \cite{llava} model and the LLM Llama-3.2-Instruct 1B \cite{llama3}.

%%%%%%%%%%%%%%%%%%%%%%%%%%%%%%
\subsection{Performance results}
We first look at the VLM LlaVa-v1.6 7B \cite{llava} model and compare~the unquantized model in FP16 resolution with the quantized versions using the lmms-eval platform \cite{lmms} on the Coco caption, VQAv2 and MMMU benchmarks. All of these benchmarks capture the ability of the model to translate visual content to text, as well as its visual and textual reasoning capability. In Table \ref{tab:table1}, we present results for our proposed QuantX quantization framework and AWQ \cite{AWQ}, which has gained a lot of traction in the industry and has been incorporated into several open source projects. It can be seen that QuantX outperforms AWQ in all metrics considered and performs within 6\% of the unquantized FP16 model. Further, this improvement in performance is achieved while using a smaller number of bits per weight (BPW) due to the larger group size in QuantX. We will update this paper with results on the run-time comparison of these algorithms in the future.
 
\begin{table*}[t]
\centering
\begin{tabular}{|p{5cm}|p{1cm}|p{2.5cm}|p{2.5cm}|p{2.3cm}|}
\hline
\textbf{Quantization configuration} & \textbf{BPW} & \textbf{Coco Caption} & \textbf{VQAv2} & \textbf{MMMU} \\ \hline

Unquantized FP16 & 16 & 1.1606 & 0.8057 & 0.343 \\ \hline

AWQ W3A16G64 & 3.25 & 1.0814 (-6.8\%) & 0.7945 (-1.4\%) & 0.328 (-4.4\%) \\ \hline

QuantX W3A16G128 & 3.14 & 1.0981 (-5.4\%) & 0.7942 (-1.4\%) & 0.332 (-3.2\%) \\ \hline

AWQ W3A8G64 & 3.25 & 1.0759 (-7.3\%)  & - & 0.327 (-4.7\%) \\ \hline

QuantX W3A8G128 & 3.14 & 1.0941 (-5.7\%) & - & 0.330 (-3.8\%) \\ \hline
\end{tabular}
\vspace{3pt}
\caption{ Comparison of the LlaVa-v1.6 model quantized using AWQ and QuantX to the unquantized model on multiple benchmarks. The QuantX framework significantly outperforms AWQ based quantization and achieves performance within 6\% of the unquantized FP16 model on all metrics.}
\label{tab:table1}
\vspace{-0.2cm}
\end{table*}

%%%%%%%%%%%%%%%%%%%%%%%%%%%%%%
\subsection{Runtime results}

\begin{figure}[b]
     \centering
     \begin{subfigure}[t]{0.49\textwidth}
         \centering
         \includegraphics[width=\textwidth]{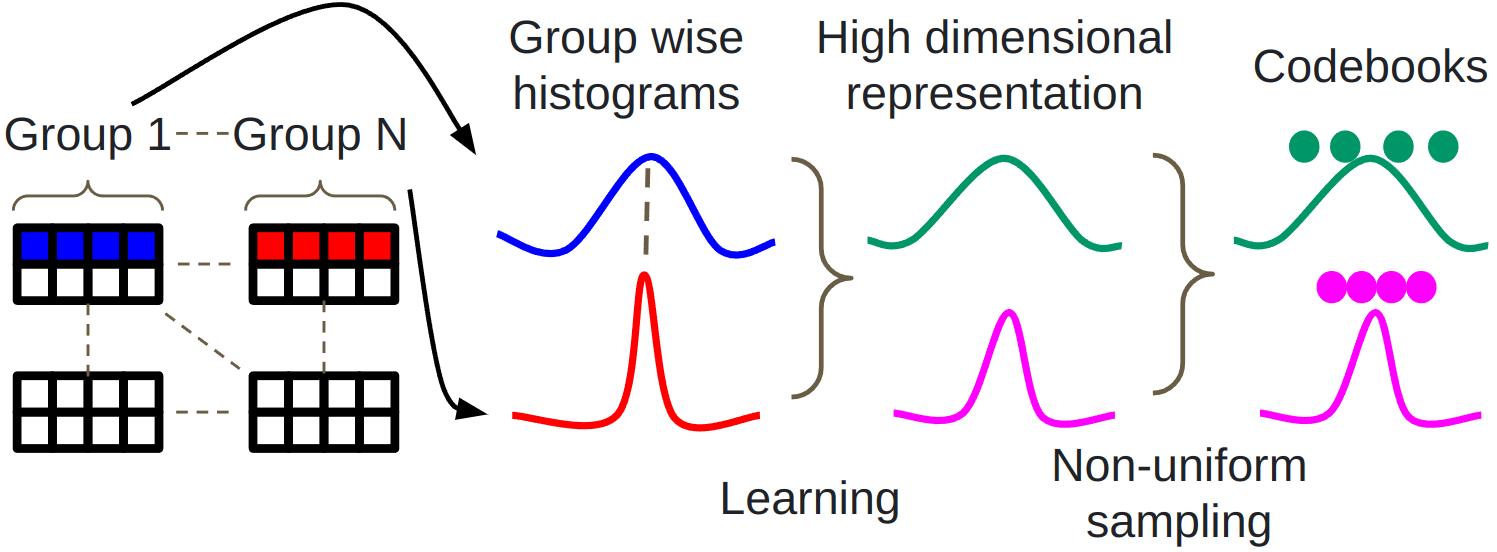}
         \caption{Functional illustration of the Q4X quantization strategy showing the histogram learning process and non-uniform quantization for dimension reduction.}
         \label{fig:Q4X}
     \end{subfigure}
     \hfill
     \begin{subfigure}[t]{0.49\textwidth}
         \centering
         \includegraphics[width=\textwidth]{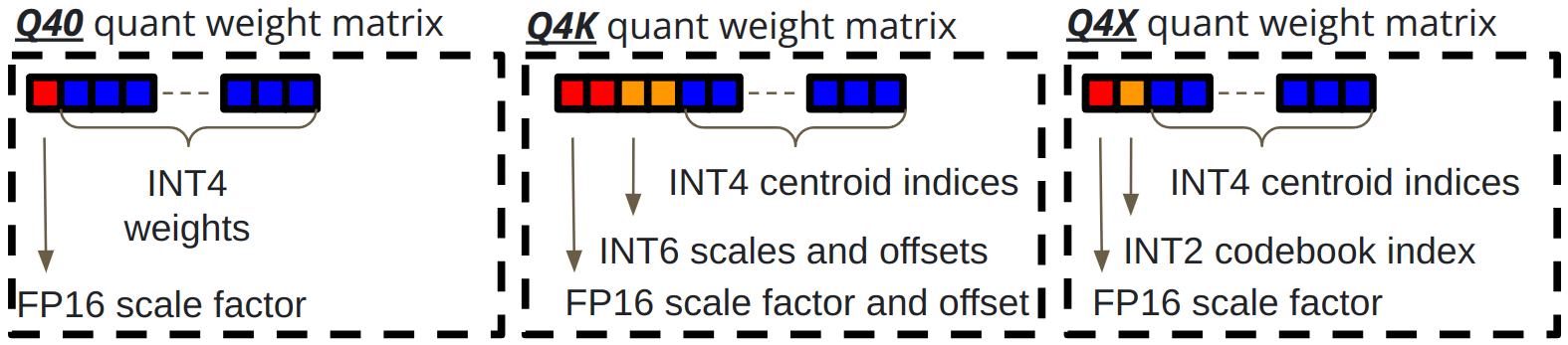}
         \caption{Illustration of the quantized weight matrices for a single group of Q40, Q4K and Q4X.}
         \label{fig:quantMatrices}
     \end{subfigure}
        \caption{ (a) Description of the proposed Q4X quantization strategy and (b) the resulting quantized matrices for Q40, Q4K and Q4X.}
        \label{fig:llamaCPP}
        \vspace{-0.0cm}
\end{figure}

Next, we demonstrate the practical feasibility of techniques in QuantX by integrating it into the popular Llama.cpp framework \cite{llamaCPP} and comparing it with the mainstream Llama.cpp techniques. Toward that end, we first give a brief overview of the built-in quantization strategies `Q40' and `Q4K' from Llama.cpp, followed by a breakdown of the new `Q4X' quantization strategy from QuantX.

\begin{itemize}

\item \textbf{Q40:} Q40 (shown on the left side of Fig. \ref{fig:quantMatrices}) is a legacy 4-bit quantization method in Llama.cpp. It works by using a rounding-to-nearest (RTN) operation on a group of 32 weights. For each group, the 4-bit weights and an FP16 scaling factor are packed into 18 bytes achieving 4.5 BPW. % Activations are dynamically quantizated in Q80, which has the exact same structure as Q40 but with 8-bit values.

\item \textbf{Q4K:} Q4K is a more recent technique introduced in Llama.cpp that uses hierarchical scaling and asymmetric quantization. Here, 256 weight elements are grouped into a 144-byte block. This block contains 256 4-bit quantized values, a 2-byte global scale and minimum, and 6-bit scales and minimums for each chunk of 32 weights. Q4K also achieves a 4.5 BPW as illustrated in the middle of Fig. \ref{fig:quantMatrices}. %Q4K uses Q8K dynamic activation quantization, which has the same structure as Q4K but with 8-bit values.

\item \textbf{Q4X:} As shown in Fig. \ref{fig:Q4X}, Q4X is a codebook-based quantization technique designed for efficient inference on CPUs. A histogram is computed for each group of 64 weights in the weight matrix followed by learning 4 histograms that best represent all the historgrams of a complete matrix. These learned histograms undergo dimension reduction, resulting in 16 non-uniform samples chosen from each. The final representation is made up of 64 centroids—16 from each of the 4 learned histograms—which are grouped into a codebook for the weight matrix. For each group of 64 weight elements, Q4X stores a 2-byte scale factor, a 2-bit codebook index that represents the best matching learned histogram for that group, and the 4-bit centroid indices for each of the 64 elements. This approach makes Q4X a 4.28 BPW (with the 2-bit codebook indices of multiple groups padded together for byte alignment) method, as illustrated on the right side of Fig.~\ref{fig:quantMatrices}. %For ease of implementation, we are using Q80 for quantizing the activations in Q4X.
\end{itemize}

The final file size, run time performance in terms of tokens per second and perplexity computed on Wikitext v2 test subset with $n_{\text{ctx}}=2048$ is shown in Table \ref{tab:table2}. Q4X outperforms Q40, achieving a higher token throughput for a reduced model size and lower perplexity score. Similarly, Q4X has a better model size and throughput than Q4K. The perplexity score, however, is slightly lower for Q4K given its use of two-level scaling for the weights and a smaller sub-group size of 32 compared to the group size of 64 used in Q4X.

\begin{table*}[t]
\centering
% \begin{tabular}{|p{5cm}|p{1cm}|p{1.5cm}|p{2.5cm}|p{2.5cm}|}
\begin{tabular}{|c|c|c|c|c|}
\hline
{\textbf{Quantization configuration}} & \textbf{BPW} & \makecell{\textbf{File size} \\ (MiB)}  & \makecell{\textbf{Prefill/Decode} \\ \textbf{throughput} \\ (tokens/sec)} & \makecell{ \textbf{Perplexity} \\ (Wikitext v2)} \\ \hline

Unquantized FP16 & 16 & $\sim$ 2300 &  0.12 / 0.06 & 13.16 \\ \hline

Q40 W4A8G32  & 4.5 & 727.75 & 6.03 / 4.28 & 15.22 \\ \hline

Q4K W4A8G32 & 4.5 & 732.25 & 7.77 / 5.06 & 14.45 \\ \hline

Q4X W4A8G64 & 4.28 & 702.38 & 8.64 / 5.29 & 14.67 \\ \hline

\end{tabular}
\vspace{3pt}
\caption{ Comparison of the file size, run time and perplexity for Llama 3.2 1B instruction tuned model and its Q40, Q4K and Q4X quantized versions running on a Milk-V Jupiter board. For a fair comparison in perplexity, we used 8-bit RTN per-token quantization for the activation tensors for all quantization techniques emulated in QuantX.}
\label{tab:table2}
\vspace{-0cm}
\end{table*}

%%%%%%%%%%%%%%%%%%%%%%%%%%%%%%%%%%%%%%%%%%%%%%%%%%%%%%%%%%%%%%%%%%%%%%%%%%%%%%%%%%%%%%%
\section{Conclusion}
In this paper, we presented insights into the LLM quantization process that motivated the various design choices made in QuantX. We demonstrated that QuantX can successfully quantize LlaVa-v1.6 to 3-bits while achieving performance within 6\% of the unquantized FP16 model. We also showed that QuantX outperformed a SOTA technique while achieving a smaller BPW ratio. Furthermore, we showed that QuantX extends the Pareto frontier across runtime, model size, and performance by integrating its 4-bit quantization into the Llama.cpp framework. We intend to progressively update this manuscript with more data and insights.

%%%%%%%%%%%%%%%%%%%%%%%%%%%%%%%%%%%%%%%%%%%%%%%%%%%%%%%%%%%%%%%%%%%%%%%%%%%%%%%%%%%%%%%
\bibliographystyle{alpha}
\bibliography{main}

\end{document}